\pdfoutput=1

\documentclass[11pt]{article}

\usepackage[]{ACL2023}
\usepackage{hyperref}
\usepackage{listings}
\usepackage{xcolor}
\usepackage{graphicx}
\usepackage{url}
\usepackage{caption}
\graphicspath{ {./images/} }
\usepackage{amsmath}
\usepackage{float}
\lstdefinelanguage{json}{
    basicstyle=\ttfamily\small,
    showstringspaces=false,
    breaklines=true,
    frame=single,
    stringstyle=\color{purple},
    keywordstyle=\color{blue}\bfseries,
    morestring=[b]",
    morekeywords={:,}, 
}

\usepackage{xfrac}

\usepackage{times}
\usepackage{latexsym}
\usepackage{graphicx}
\graphicspath{{./images/}}

\usepackage[T1]{fontenc}

\usepackage[utf8]{inputenc}

\usepackage{microtype}

\usepackage{inconsolata}

\title{ComicScene154: A Scene Dataset for Comic Analysis}

\author{Sandro Paval, Ivan P. Yamshchikov, Pascal Meißner}

\begin{document}
\maketitle
\begin{abstract}
Comics offer a compelling yet under-explored domain for computational narrative analysis, combining text and imagery in ways distinct from purely textual or audiovisual media. We introduce \textit{ComicScene154}, a manually annotated dataset of scene-level narrative arcs derived from public-domain comic books spanning diverse genres. By conceptualizing comics as an abstraction for narrative-driven, multimodal data, we highlight their potential to inform broader research on multi-modal storytelling. To demonstrate the utility of \textit{ComicScene154}, we present a scene segmentation baseline, providing an initial benchmark for future studies to build upon. Our results indicate that \textit{ComicScene154} constitutes a valuable resource for advancing computational methods in multimodal narrative understanding and expanding the scope of comic analysis within the Natural Language Processing community.
\end{abstract}

\section{Introduction}

Advancements in Computer Vision (CV) and Natural Language Processing (NLP) have enabled the analysis and processing of diverse media types in both standalone and multimodal settings \cite{bayoudh2022survey}. However, one medium that remains underexplored is comics, which uniquely interweaves text and images into panels, creating a sequence of discontinuous frames. The narrative link between these frames is implicit, requiring a cognitive process called \textit{closure} to bridge the gaps. This characteristic makes comics particularly intriguing not only as a distinct medium, but also as a compact abstraction for more continuous, multimodal data such as videos or movies \cite{Rao_2020_CVPR, cohn2013visual}. By sampling frames from data streams and compiling them in a comic-like structure, one can preserve narrative structure while reducing the data’s overall complexity.

Progress in computational comic analysis largely falls into four categories: \textit{object detection tasks}~\cite{ogawa2018object,yanagisawa2018study}, \textit{multi-modal understanding}~\cite{li2024zero,rigaud2024toward}, \textit{image generation}~\cite{proven2021comicgan,yang2021automatic,wang2012movie2comics} and \textit{dataset proposals} \cite{iyyer2017amazing,dunst2017graphic}. In comparison, the problem of \textit{narrative understanding} \cite{pratt2009narrative} in comics remains relatively underdeveloped.

Comic analysis tasks, such as character re-identification \cite{sachdeva2024manga} and closure inference \cite{iyyer2017amazing}, would rely on narrative context but have to resort to operating at broad levels (e.g., entire comics) or strict physical divisions (e.g., single pages). While page-wise processing may be practical from a layout point-of-view, the internal narrative structure is more nuanced \cite{zehe2021detecting,cohn2010limits}. Segmenting comics according to narrative arcs could better support tasks like story summarization \cite{huang2016visual} and entity tracking \cite{kim2023entity} by ensuring that only relevant information is included. Conversely, using an entire comic might introduce superfluous content, whereas strict physical divisions risk omitting key story details.

This take on segmentation could also serve as a simplified approach to scene segmentation in video data~\cite{Rao_2020_CVPR}, where select frames are extracted to create a comic-like structure conducive to smaller-scale analysis. Beyond comptutational problems, our take on segmentation could benefit tasks in digital humanities: identifying recurring storytelling patterns, comparing artistic styles, or performing linguistic analyses on segmented narrative arcs \cite{cohn2013visual,zehe2021detecting}.

\begin{table*}[h]
\centering
\begin{tabular}{l l l l l l l l}
\hline
\textbf{Dataset}  & \textbf{Tasks} & \textbf{Years} & \textbf{Style} & \textbf{Books} & \textbf{Pages} \\
\hline
eBDtheque & d,t & 1905-2012 & mix & 28 & 100 \\
COMICS & c & 1938-1954 & comics & 3948 & 198000 \\
GCN & d,t & 1978-2013 & comics & 253 & 38000 \\
DCM772 & d & 1938-1954 & comics & 27 & 772 \\
Manga109 & d,t,r & 1970-2010 & manga & 109 & 10000 \\
BCBId & - & - & bangla & 64 & 3000 \\
PopManga & d,t,r & 2010-2023 & manga & 25 & 1800 \\
CoMix & d,t,r,N,D  & 1938-2023 & mix & 100 & 3800 \\
\hline
\textbf{Ours} & d,S & 1942-1962 & comics & 4 & 154 \\
\hline
\end{tabular}
\caption{Overview of comic datasets including ours, adapted from \cite{vivolicomix}. \textbf{Tasks:} Classification (c), Detection (d), Text-Character Association (t), Character Re-Identification (r), Character Naming (N), Dialog (D), and Scene Segmentation (S).}
\label{tab:extern_datasets}
\end{table*}

In this paper, we aim to advance all these fields by introducing the scene-annotated dataset for comics \textit{ComicScene154}. We devised a prototypical approach to display the limitations and challenges associated with such a task. This dataset can facilitate new segmentation methods while providing a benchmark for existing approaches. Additionally, it offers a consistent resource for tasks such as story summarization and character identification on a more fine-grained, narrative-based scale.
Section 3 will detail the dataset construction and the evaluation, while Section 4 will present an example to illustrate the inherent challenges of the task.

\section{Related Work}
\subsection{Scenes in Comics and Beyond} \label{sec:movie}

We follow the definition of scenes from \cite{Rao_2020_CVPR}, and describe a scene as a plot-based semantic unit, in which an overarching task is pursued by a certain cast of characters. In most cases, these semantic units maintain temporal and spatial coherence. Additionally, we treat a scene analogous to narrative arcs as defined by \cite{cohn2013visual}

In movies, different shots—sequences of images captured from the same viewpoint—form the visual narrative. Similarly, comics consists of panels that depict imagery from diverse viewpoints. By selecting a representative image from each shot, along with its corresponding narrative data (e.g., spoken text or script), the resulting sequence can be compacted. By recreating a comic-like structure via representative frames, image data can be significantly reduced beyond comics.

\subsection{Analogy to Semantic Text Segmentation}
In NLP, semantic text segmentation involves breaking text down into semantically coherent segments or representations, often focusing on higher-level logical forms\footnote{For a detailed survey on semantic text segmentation, we refer the reader to \cite{kamath2019survey}}. This task shares conceptual parallels with scene segmentation in comics: both seek to identify coherent boundaries within a data stream - be it textual sentences or visual-narrative frames.

To evaluate segmentation quality, text-based approaches often rely on the $p_k$ metric, which quantifies the proportion of sentences (or segments) that are incorrectly “cut” or “joined” \cite{glavavs2016unsupervised} via segmentation. We adopt a similar perspective for evaluating scene segmentation: just as $p_k$ measures segmentation consistency in text, we devise and adapt an analogous metric for assessing how effectively consecutive frames (or panels) are grouped into scenes within comics and movies.

\subsection{Comic Datasets}\label{ref:datasets}
Obtaining ethically sourced data for comics is non-trivial due to the commercial nature of the medium. The diverse datasets for manga—most notably \textit{Manga109} \cite{fujimoto2016manga109} and \textit{PopManga}\cite{sachdeva2024manga}—are not a suitable replacement, as mangas often differ considerably from classic Western comics in terms of storytelling conventions, structural layout, and linguistic features \cite{cohn2011different}. These differences highlight the need for building datasets tailored to distinct comic traditions, enabling more targeted research in both CV and multimodal NLP tasks.

Western datasets on the other hand consist almost exclusively of public domain comics, which restricts them to older sources. Notable datasets compiled in Table \ref{tab:extern_datasets} include DCM772\cite{nguyen2018digital}, focusing on object detection and association, or the combined manga and comic dataset CoMix\cite{sachdeva2024manga} which draws from already existing datasets for benchmarking purposes.  In terms of cultural diversity, eBDtheque \cite{guerin2013ebdtheque}, offers not only a mix of manga and American comics, but also French bande dessinée, while BCBId\cite{dutta2022bcbid} offers Bangladeshi comics.

\section{The ComicScene154 Dataset}

\subsection{Data Source}

For the sake of reproducibility, only freely available public-domain data were used in this study. While large-scale datasets exist for Japanese manga, we opted not to include them here due to the format discrepancies discussed in Section~\ref{ref:datasets}. All data was collected from \textit{Comic Book Plus}\footnote{https://comicbookplus.com}, a repository offering a diverse range of public-domain comics. The comics used skew toward older storytelling and artistic styles, reflecting their origins in the “Golden Age” of comics, roughly 1940–1960. 

\begin{table}[t]
\small
\begin{tabular}{llllll}
\hline
\textbf{Comic, Volume}& \textbf{Pages} & \textbf{Panels}& \textbf{Scenes} & \textbf{Genre}\\
\hline
\text{Alley Oop,1} & \text{35} & \text{191} & \text{37} & \text{Humor} \\
\text{Champ Comics,24} & \text{38} & \text{263} & \text{55} & \text{Heroes} \\
\text{Treasure Comics,6}  & \text{41} & \text{221} & \text{41} & \text{Fantasy}\\
\text{Western Love,4}  & \text{40} & \text{279} & \text{52}  & \text{Love} \\
\hline
\end{tabular}
\caption{Details on of our \textit{ComicScene154} dataset.}
\label{tab:dataset}
\end{table}

\subsection{Dataset}
Our dataset, \textit{ComicScene154}, comprises four public-domain comic magazines containing 34 distinct stories that span a total of 154 pages across various genres and publication years\footnote{Here is the link to the dataset and code \url{https://github.com/Knorrsche/ComicScene154}}. These magazines were selected to ensure diversity in storytelling, thus covering multiple narrative styles. Table~\ref{tab:dataset} outlines the chosen comics and their associated metadata. Due to the focus on narrative, none of the existing datasets in Table \ref{tab:extern_datasets} were used, as their data is not categorized by genre or consist of samples of comics. The latter is particularly important because being able to segment the dataset at the level of full stories is crucial, as scene construction relies on the broader comic narrative. Randomly sampled panels or pages are therefore not suitable for the task of scene segmentation, we address.

\begin{figure}
    \centering
    \includegraphics[width=7cm,height=5cm]{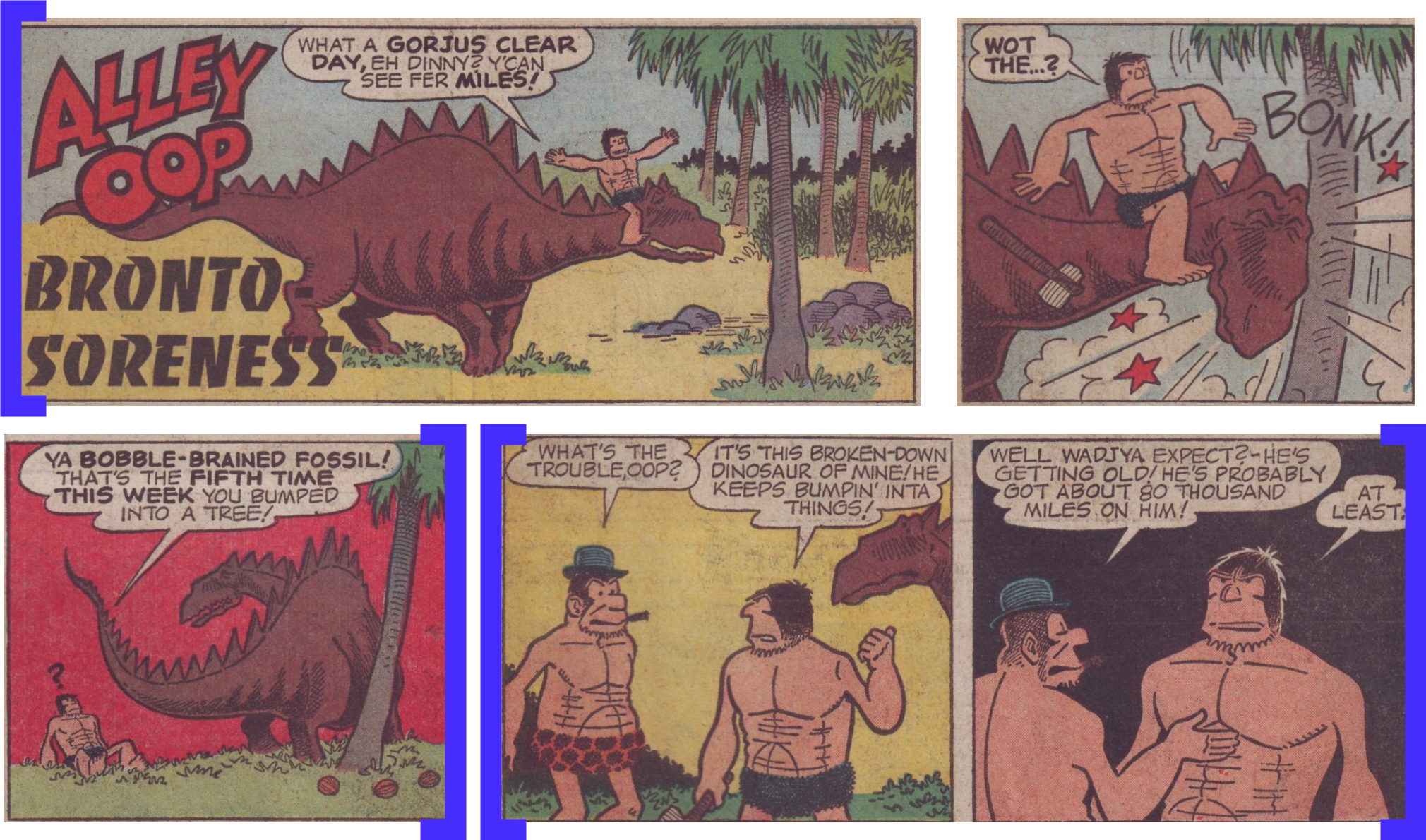}
    \caption{Panels with 2 scenes visualized as intervals.}
    \label{fig:Scene_example}
\end{figure}

Before distributing the dataset to the annotators, all panels were extracted from comic pages along with their coordinates, and then numbered in reading order. Annotating suchlike ensures the consistency of the dataset, as any discrepancies in panel numbering (e.g., missing or extra panels) would corrupt the reading order. Additionally, each panel was tagged with a Boolean value indicating whether it marks the start of a new scene. An example for visualization can be seen at Figure \ref{fig:Scene_example}, where the scene boundaries are marked blue.

\subsection{Reliability}
Given the inherently subjective nature of scene segmentation, evaluating the reliability of our annotations is critical. To this end, one-third of the dataset was independently labeled by three different groups of two annotators  (Tester 1 \& Tester 2), and their annotations were compared against our own. This triple annotation process not only ensures data consistency but also gauges whether different annotators share a conceptual understanding of scene boundaries. Annotators were tasked with marking each panel that signals the start of a new narrative arc. Following the definition in Section~\ref{sec:movie}, they received a brief introduction to the task and a sample annotated example for guidance.

To quantify agreement, we employed the $p_k$ metric from the semantic text segmentation literature \cite{glavavs2016unsupervised}. This measure compares two segmentations by assessing the proportion of sliding windows that are inconsistently segmented. In our context, $p_k = 0$ indicates perfect alignment of scene boundaries, while $p_k = 1$  indicates total misalignment. The choice of window size $k$ is typically informed by half of the average length of segments. Lacking an external reference for segment length, we computed $k$  by averaging the scene lengths from our own annotations and those produced by the other annotators, yielding $k = 3$.

\begin{table}[h]
    \centering
    \small
    \begin{tabular}{l|c|c|c}
        \hline
        \textbf{Excerpts} & \textbf{Tester 1} & \textbf{Tester 2} & \textbf{In-between} \\
        \hline
        Alley Oop (1) &  0.07 & 0.00 & 0.07 \\
        Champs (1) & 0.27 & 0.16 &  0.22\\
        Champs (2) & 0.21 & 0.07 & 0.28 \\
        Treasure C. (1) & 0.06 & 0.19 & 0.14 \\
        Treasure C. (2) & 0.17 & 0.33 & 0.26 \\
        Western L. (1) & 0.12 & 0.37 & 0.29\\
        \hline
        \textbf{Average} & 0.15 & 0.19 & 0.21 \\
        \hline
    \end{tabular}
    \caption{Agreement scores (\( p_k \)) for 6 randomly chosen excerpts. Each excerpt was tested by 2 of 6 tester.}
    \label{tab:agreement_scores}
\end{table}

With an average tester \( p_k \) score of \( 0.17 = \sfrac{0.15+0.19}{2} \), there is some notable agreement in the interpretation of what a scene is yet as with other narrative related tasks there is certain noise associated with the subjectivity of a given assessor. This suggests that though the overall concepts align reasonably well in Table \ref{tab:agreement_scores}, a major challenge in scene segmentation remains the lack of an intersubjective definition of a scene. The varying scores illustrate this issue—most notably in the excerpt extracted from \textbf{Western Love}, where Tester 2 exhibited the highest disagreement with a score of 0.37.  
When analyzing the annotation data, it became clear that Tester 2 segmented the comic into much shorter scenes. This behavior can be attributed to the inherent subjectivity of the interpretation of narratives.

\section{Scene Segmentation Benchmark}
For benchmarking the dataset, a two step scene segmentation pipeline was developed:
First, utilizing a multi-modal model to predict scene boundaries and then refining the predictions using a reasoning-based large language model (LLM).

For both steps, Gemini's reasoning model, \texttt{gemini-2.0-flash-thinking-exp}, was used. In the initial iteration, each comic page, along with the coordinates of the panels and their reading order, was provided as context alongside a prompt. The model's task was to generate descriptions of narrative arcs and identify the panel where each arc begins, and hence scenes. While these outputs were useful, their quality remained limited.

Due to the non-deterministic nature of LLM, the performance was evaluated using the $p_k$ metric, based on ten iterations per comic. Table~\ref{tab:agreement_scores_multi} presents the average scores in these ten iterations, compared to scenes we defined randomly. The results indicate that the performance of the multi-modal model is only marginally better than random definition, highlighting the current limitations. Additionally, the agreement between the ten iterations was analyzed, revealing high consistency. This suggests that while the model struggles to fully align with human-annotated scenes, it does capture some underlying patterns. It also highlights the present challenge of subjectiveness in scene segmentation.

\begin{table}[H]
    \centering
    \small
    \begin{tabular}{l|c|c|c}
        \hline
        \textbf{Comic} & \textbf{Dataset}  & \textbf{Random} & \textbf{In-between} \\
        \hline
        Alley Oop &  0.43 & 0.47 & 0.10  \\ 
        Champs & 0.40 & 0.43 & 0.09 \\
        Treasure Comics & 0.43 & 0.46 & 0.05 \\
        Western Love & 0.43 & 0.46 & 0.07 \\
        \hline
        \textbf{Average} & 0.42 & 0.46 & 0.06\\
        \hline
    \end{tabular}
    \caption{Agreement scores (\( p_k \)) of multi-modal, random and in-between different multi-modal iterations}
    \label{tab:agreement_scores_multi}
\end{table}

To further refine the results, the initial model outputs were processed using a reasoning-based LLM. Since this refinement step depends on the output of the multi-modal model, each of the ten iterations was further benchmarked ten times, resulting in a total of 100 evaluations per comic. Table~\ref{tab:agreement_scores_refined} compares the average score across all 100 iterations to the best average score from a single multi-modal output (based on ten iterations). 

While the overall output still exhibits a relatively high $p_k$ score, and the improvement from averaging 100 iterations is marginal, the best-performing iterations for each comic show noticeable gains. Although these best results represent the most favorable outcomes, it is important to note that they still consist of ten independent iterations based on the same multi-modal model output. These large performance variations further illustrate the inherent challenges and subjectivity in scene segmentation.

\begin{table}[H]
    \centering
    \small
    \begin{tabular}{l|c|c|c}
        \hline
        \textbf{Comic} & \textbf{All Iter.} & \textbf{Best Iter.}  & \textbf{In-between} \\
        \hline
        Alley Oop &  0.36  &  0.27 & 0.04 \\ 
        Champs & 0.40  & 0.37 & 0.05\\
        Treasure Comics & 0.41  & 0.34 & 0.05 \\
        Western Love & 0.39 & 0.36 & 0.04 \\
        \hline
        \textbf{Average} & 0.39 &  0.34  & 0.05\\
        \hline
    \end{tabular}
    \caption{Refined agreement scores (\( p_k \))}
    \label{tab:agreement_scores_refined}
\end{table}

\section{Discussion}

Comparing the results of both human (Table \ref{tab:agreement_scores}) and AI (Tables \ref{tab:agreement_scores_multi}, \ref{tab:agreement_scores_refined}) benchmarks reveals a significant difference. While the human benchmark showed notable improvement compared to randomly defined scenes, the multi-modal approach demonstrated almost no improvement. The refined method did improve results, but the largest performance gains were observed only in specific iterations rather than across all outputs. This highlights the present challenge of defining scenes, as even human annotators disagree. Nevertheless, while the refined performance relied on the multi-modal output, the annotations themselves showed low differences in between, demonstrating the model's ability to detect certain patterns—though not the semantic units we aimed to capture, i.e., the scenes.

\section{Conclusion}

We introduce ComicScene154, a novel dataset designed to facilitate the study of scene segmentation in comics. By addressing the unique challenges (e.g. closure, narrative segmentation) of this medium, ComicScene154 provides a robust foundation for developing algorithms capable of understanding complex narrative structures like scenes (narrative arcs). Furthermore we propose the leveraging of comic analysis techniques on formats other than comics, like movies, to reduce the data scale and complexity of tasks such as narrative summarizations or object clustering.

\section*{Limitations}

Despite its contributions, ComicScene154 has certain limitations. The primary challenge remains the subjectivity of scene segmentation, as interpretations can vary across annotators. This was also seen in the performance of scene segmentation, where the average annotations of the models displayed lacking results. Additionally, the dataset is largely composed of "Golden Age" comics, which may limit its applicability to modern comics due to differences in artistic style, storytelling techniques, and narrative complexity.

\section*{Ethics Statement}

This paper complies with the \href{https://www.aclweb.org/portal/content/acl-code-ethics}{ACL Ethics Policy}. The comics included in ComicScene154 are in the public domain, ensuring compliance with copyright regulations. However, we acknowledge the potential for annotator bias, which could impact dataset consistency. To mitigate this, we implemented a standardized annotation framework and conducted inter-annotator agreement evaluations to enhance reliability.

\section*{Acknowledgments}

We extend our gratitude to the annotators for their meticulous work and to the reviewers for their valuable feedback.

\bibliographystyle{acl_natbib}
\bibliography{anthology,custom}

\section{Appendix}
\subsection{Annotators}

The annotators were friends or colleagues of the authors who followed the provided guidelines without any additional input from the authors. They volunteered without compensation. 

\subsection{Guideline}
\noindent \textbf{Dear Participant,}

I warmly invite you to participate in a survey designed to evaluate the accuracy of a dataset I have developed. The focus of this survey is to assess how effectively the dataset can identify the start and end points of scenes in comics.

\textbf{Purpose of the Survey:}

The segmentation of comic scenes is often subjective and can vary from reader to reader. Sometimes, it is difficult to pinpoint exactly where one scene ends and the next begins. In such cases, deviations are completely acceptable. What is most important is whether there is significant agreement on the core panels of a scene—the key panels that form the heart of the storyline.

\textbf{Instructions:}

\begin{enumerate}
    \item \textbf{Review the comic images:} You will be provided with a selection of comic images representing different scenes.
    \item \textbf{Mark the start and end points:} Indicate or describe where you perceive the transition from one scene to the next.
    \item \textbf{Reference example:} A sample comic is attached for your orientation, but please note that it is merely a reference—you do not need to strictly follow it.
\end{enumerate}

\textbf{Important Note:}

Comics are a visual art form, and often, there is no definitive answer to where a scene begins or ends. Your feedback will help me identify patterns and commonalities that will be valuable in improving the dataset.

\subsection{Additional Statistics}
\begin{figure}[H]
    \centering
    \includegraphics[width=\textwidth]{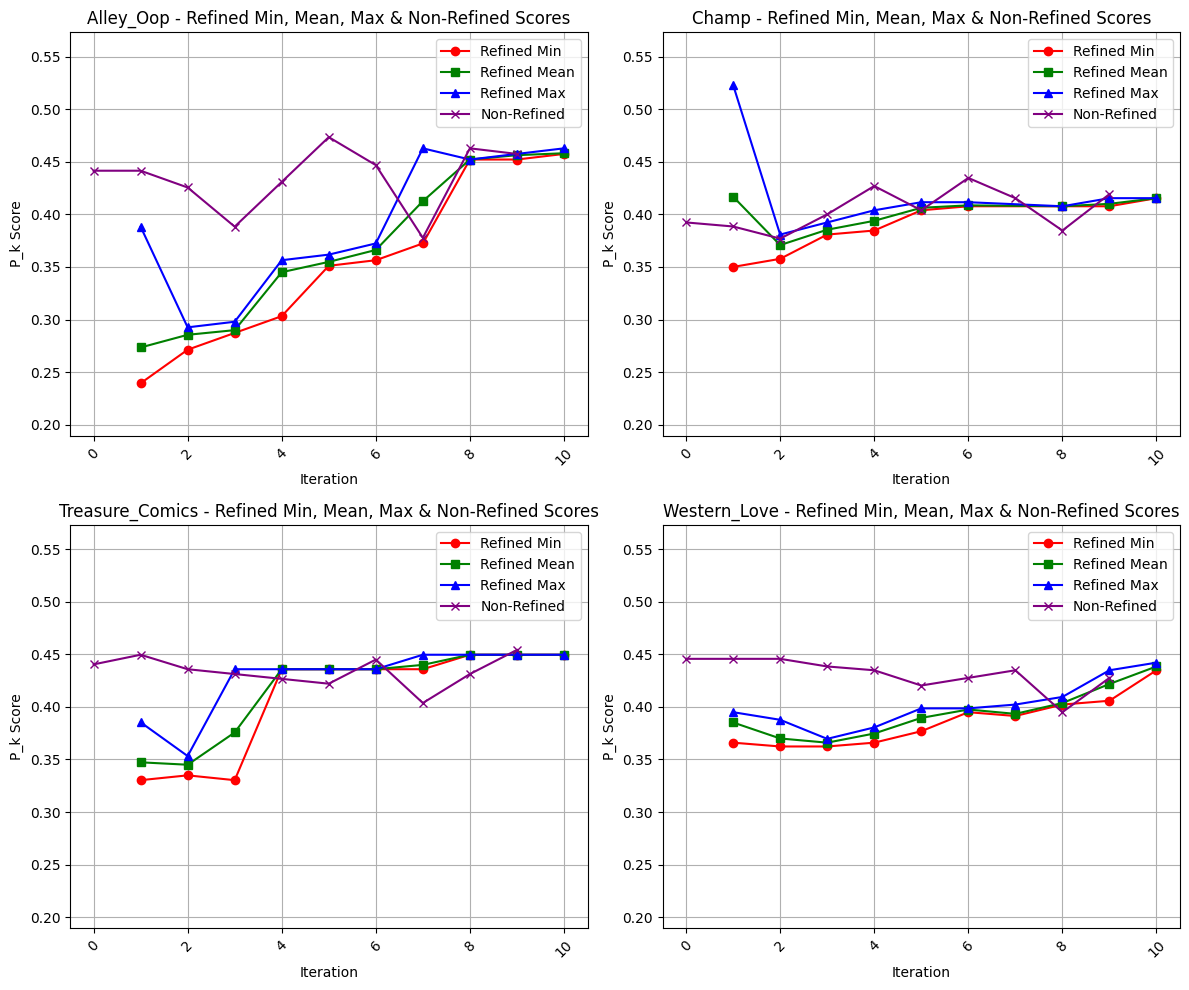}
\end{figure}

\end{document}